\documentclass[10pt,journal]{IEEEtran}

\usepackage{amsmath,amssymb}
\usepackage{booktabs}
\usepackage{tabularx}
\usepackage{array}
\usepackage{multirow}
\usepackage{url}
\usepackage{xcolor}
\usepackage{cite}
\usepackage{algorithm}
\usepackage{algorithmic}
\usepackage{graphicx}
\usepackage{enumerate}
\usepackage{comment}

\newcommand{\RPR}{\mathrm{RPR}}
\newcommand{\Fone}{\mathrm{F1}}
\newcommand{\Score}{\mathrm{Score}}
\newcommand{\NGM}{N_{\mathrm{GM}}}
\newcommand{\Nsol}{N_{\mathrm{sol}}}
\newcommand{\Aset}{\mathcal{A}}
\newcommand{\Cset}{\mathcal{C}}
\newcommand{\Pset}{\mathcal{P}}
\newcommand{\Rset}{\mathcal{R}}
\newcommand{\Mset}{\mathcal{M}}
\newcommand{\Dnorm}{d_{\mathrm{norm}}}
\graphicspath{{images/}}

\title{S-CARD-CMSA: A Score-Aware Candidate Archive with Density-Filtered Reporting for Multimodal Optimization}

\author{\IEEEauthorblockN{Dikshit Chauhan}\\
\IEEEauthorblockA{Department of Electrical and Computer Engineering,\\ National University of Singapore}
\thanks{This manuscript describes the method prepared for the IEEE CEC 2026 Competition on Benchmarking Niching Methods for Multimodal Optimization. \texttt{Email: dikshitchauhan608@gmail.com}}}

\begin{document}
\maketitle

\begin{abstract}
Multimodal optimization aims to locate multiple globally optimal or near-optimal solutions in a single run. This paper presents \emph{S-CARD-CMSA}, a score-aware candidate-archive and density-filtered reporting framework built on the covariance matrix self-adaptation evolution strategy with repelling subpopulations (RS-CMSA-ESII). The method is developed for the IEEE CEC 2026 Competition on Benchmarking Niching Methods for Multimodal Optimization. Rather than modifying the core search dynamics of RS-CMSA-ESII, S-CARD-CMSA preserves its sampling, covariance adaptation, taboo-region update, restart, and termination mechanisms. Two conservative extensions are introduced. First, a passive secondary candidate archive records the restart-level best candidates without influencing the search trajectory. Second, a score-aware density-filtered reporting rule constructs the final solution set by balancing robust peak ratio and precision-driven F1-score. Development experiments show that the density-filtered rule preserves the peak coverage obtained by a medium score-aware rule while reducing redundant reports. On a broader validation subset, it maintains the same mean RPR while improving mean precision, F1-score, and the official-score-oriented average. The method does not use true global-minimum locations during optimization; such information is used only for offline development analysis and post-run scoring. The source code of S-CARD-CMSA is available at \url{https://github.com/ChauhanDikshit}.
\end{abstract}


\section{Introduction}
Many real-world optimization problems possess more than one high-quality solution. In engineering design, decision support, scientific modeling, and planning problems, several distinct solutions with comparable objective values may be equally useful because they provide choices under unmodeled preferences, robustness requirements, or implementation constraints \cite{chauhan2026review}. This motivates multimodal optimization (MMO), where the goal is not only to locate one global optimum but also to identify multiple distinct global minimizers in a single run \cite{li2017survey,chauhan2025advancements}. Compared with conventional single-solution optimization, MMO requires an algorithm to simultaneously maintain search diversity, converge accurately to promising basins, and avoid reporting multiple redundant solutions corresponding to the same optimum.

Niching and archive-based strategies have been widely studied for MMO problems. Classical niching methods include fitness sharing, crowding, clearing, and speciation, while more recent approaches use neighborhood structures, clustering, multi-swarm search, hill-valley tests, landscape awareness, and covariance adaptation. A representative comparison of relevant MMO methods is summarized in Table~\ref{tab:related_methods}. Among covariance-adaptive methods, RS-CMSA-ES \cite{ahrari2017rscmsa} and its improved version RS-CMSA-ESII \cite{ahrari2022rscmsa2} are particularly relevant because they combine covariance matrix self-adaptation with archive-based repelling mechanisms and restart-based search. RS-CMSA-ESII improved several components of the original RS-CMSA-ES, including taboo-distance adaptation, covariance adaptation with elite solutions, termination criteria, bound handling, and initialization efficiency~\cite{ahrari2022rscmsa2}. Owing to these features, RS-CMSA-ESII represents a strong base optimizer for modern MMO benchmarks. It has shown competitive performance against previous winners of niching competitions, including NEA2 \cite{preuss2012nea2}, NMMSO \cite{fieldsend2014nmmso}, and HillVallEA~\cite{maree2019hillvallea}.

\begin{table*}[t]
\centering
\caption{Representative MMO methods and their relevance to the present work.}
\label{tab:related_methods}
\begin{tabularx}{\textwidth}{p{2cm}p{3cm}X p{4.1cm}}
\toprule
Method family & Representative methods & Main idea & Relevance to this work \\
\midrule
Classical niching & Fitness sharing, crowding, clearing, speciation & Maintain diversity by penalizing crowded regions, restricting replacement, or forming species around promising individuals & Establishes the basic principle that multiple basins must be preserved during optimization. \\
\midrule
Neighborhood-based methods & Local/neighborhood DE, LIPS, CNMM & Restrict information exchange or variation operators to nearby individuals & Useful for preserving local basins, but performance can depend strongly on distance measures and neighborhood definitions. \\
\midrule
Clustering-based niching & NBC/NEA2, APC-based methods, DBSCAN-based methods & Partition the population into clusters or species that approximate different basins & Motivates the use of candidate grouping and duplicate control in the final reporting stage. \\
\midrule
Multi-swarm methods & NMMSO and related multiswarm approaches & Maintain multiple swarms that can split, merge, or migrate to cover different optima & Effective for explicit basin coverage, but may require careful control of swarm creation, merging, and population allocation. \\
\midrule
Hill-valley-based methods & HillVallEA and related approaches & Use hill-valley tests to determine whether two solutions belong to the same basin & Closely related to basin verification and archive update mechanisms used in RS-CMSA-ESII. \\
\midrule
Covariance-adaptive niching & RS-CMSA-ES, RS-CMSA-ESII & Use covariance-adaptive local search together with repelling taboo regions around archived optima & Selected as the base optimizer because it can adapt to basin shape and supports archive-based repulsion. \\
\midrule
Proposed method & S-CARD-CMSA & Preserve the RS-CMSA-ESII search engine; add passive secondary candidate retention and score-aware density-filtered final reporting & Targets the CEC 2026 RPR-F1 scoring trade-off by improving final candidate selection without disrupting the robust core search. \\
\bottomrule
\end{tabularx}
\end{table*}
The IEEE CEC 2026 Competition on Benchmarking Niching Methods for Multimodal Optimization\footnote{\url{https://sites.google.com/view/tf-mmo/activities/competitions/mmocec2026}} introduces an updated benchmark suite designed to evaluate MMO algorithms under scalable and structurally diverse landscapes. The suite consists of 16 problem identifiers, 15 instances per problem, and four dimensions, $D\in\{2,5,10,20\}$, resulting in 960 official optimization cases~\cite{ahrari2026tr}. The benchmark includes basins with different shapes, sizes, orientations, and variable-dependency structures. The evaluation budget is fixed at $FE_{\max}=20,000D.$ Therefore, an effective method must use the available function evaluations carefully to discover multiple basins while maintaining sufficient local accuracy.

A key aspect of the CEC 2026 assessment is that the final score is not determined by peak coverage alone. The score combines robust peak ratio (RPR) and F1-score. While RPR measures the fraction of global minimizers successfully identified, the F1-score also depends on precision, which is affected by the number of reported solutions. Consequently, reporting too many redundant or low-quality candidates can reduce the final score even when RPR is high. This creates a practical trade-off: the algorithm should report enough candidates to cover distinct global minimizers, but it should avoid excessive reporting that harms precision.

This observation motivates the present work. RS-CMSA-ESII already contains carefully designed mechanisms for sampling, covariance adaptation, taboo-region construction, restart control, and archive update. Directly modifying these core mechanisms may disturb the balance between exploration and exploitation. Instead, this work proposes a conservative extension that preserves the RS-CMSA-ESII search engine and improves the candidate-retention and final-reporting stages. The proposed method is named \emph{S-CARD-CMSA}, where S-CARD denotes Score-aware Candidate Archive with Density-filtered Reporting, and CMSA indicates that the framework is built on RS-CMSA-ESII. The method introduces a passive secondary candidate archive to record useful restart-level candidates and a score-aware density-filtered rule to construct the final submitted solution set. The main contributions of this work are summarized as follows.
\begin{itemize}
\item A score-aware candidate-retention and reporting framework is proposed for the CEC 2026 MMO benchmark.
\item A passive secondary candidate archive is introduced to recover promising restart-level candidates that were visited but not retained by the primary archive.
\item A density-filtered final reporting rule is developed to reduce redundant reported candidates while preserving peak coverage. This directly targets the RPR-F1 trade-off used in the CEC 2026 scoring rule.
\item A staged experimental development process is reported. The final method is compared with the original primary-archive baseline and several intermediate variants.
\end{itemize}

The remainder of this paper is organized as follows. Section~\ref{sec:background} reviews the CEC 2026 scoring rule and the base RS-CMSA-ESII algorithm. Section~\ref{sec:proposed} presents the proposed S-CARD-CMSA framework. Section~\ref{sec:experiments} describes the experimental development process, ablation variants, and validation results. Section~\ref{sec:discussion} discusses the main observations, limitations, and future research directions. Finally, Section~\ref{sec:conclusion} concludes the paper.

\section{Background}\label{sec:background}
\subsection{CEC 2026 Multimodal Optimization Benchmark}
The IEEE CEC 2026 Competition on Benchmarking Niching Methods for MMO evaluates algorithms on a scalable multimodal benchmark suite. The benchmark consists of 16 problem identifiers, 15 instances for each problem, and four dimensions, $D\in\{2,5,10,20\}$, which results in $16\times 15\times 4 = 960$ official optimization cases~\cite{ahrari2026tr}. For each case, the maximum number of objective-function evaluations is defined as $FE_{\max}=20,000D.$

\begin{figure*}[t]
\centering
\includegraphics[width=0.24\textwidth]{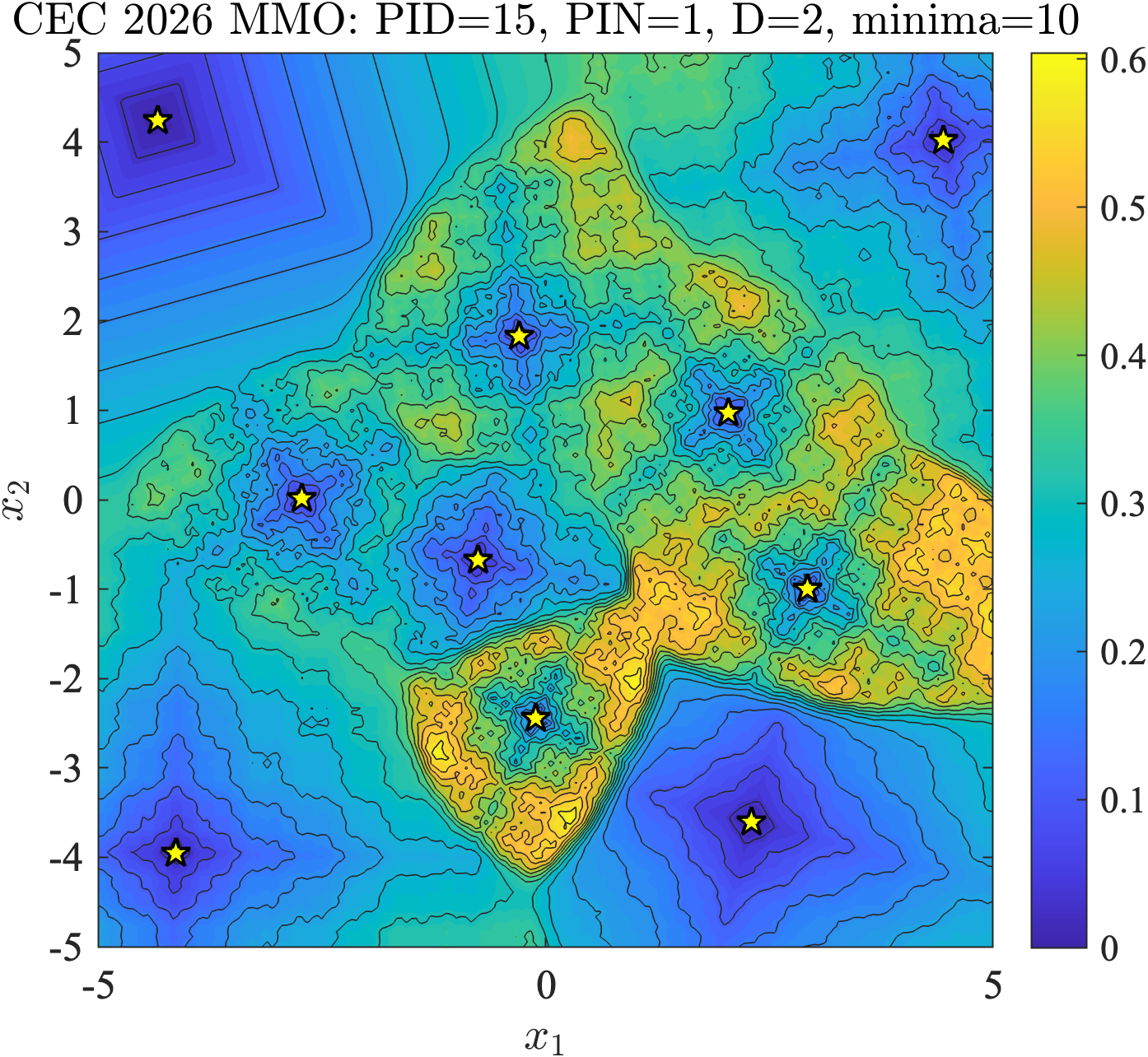}
\includegraphics[width=0.24\textwidth]{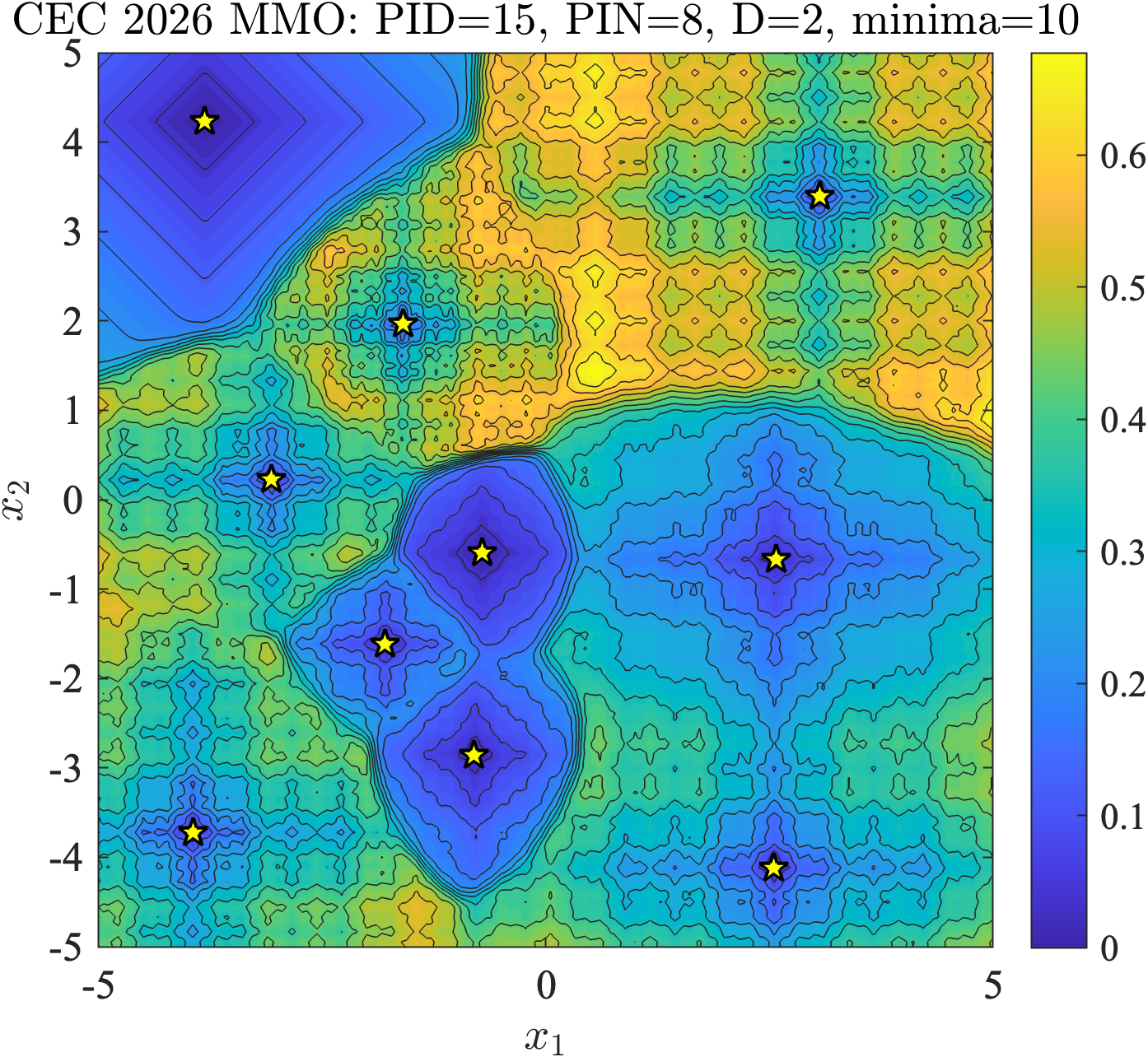}
\includegraphics[width=0.24\textwidth]{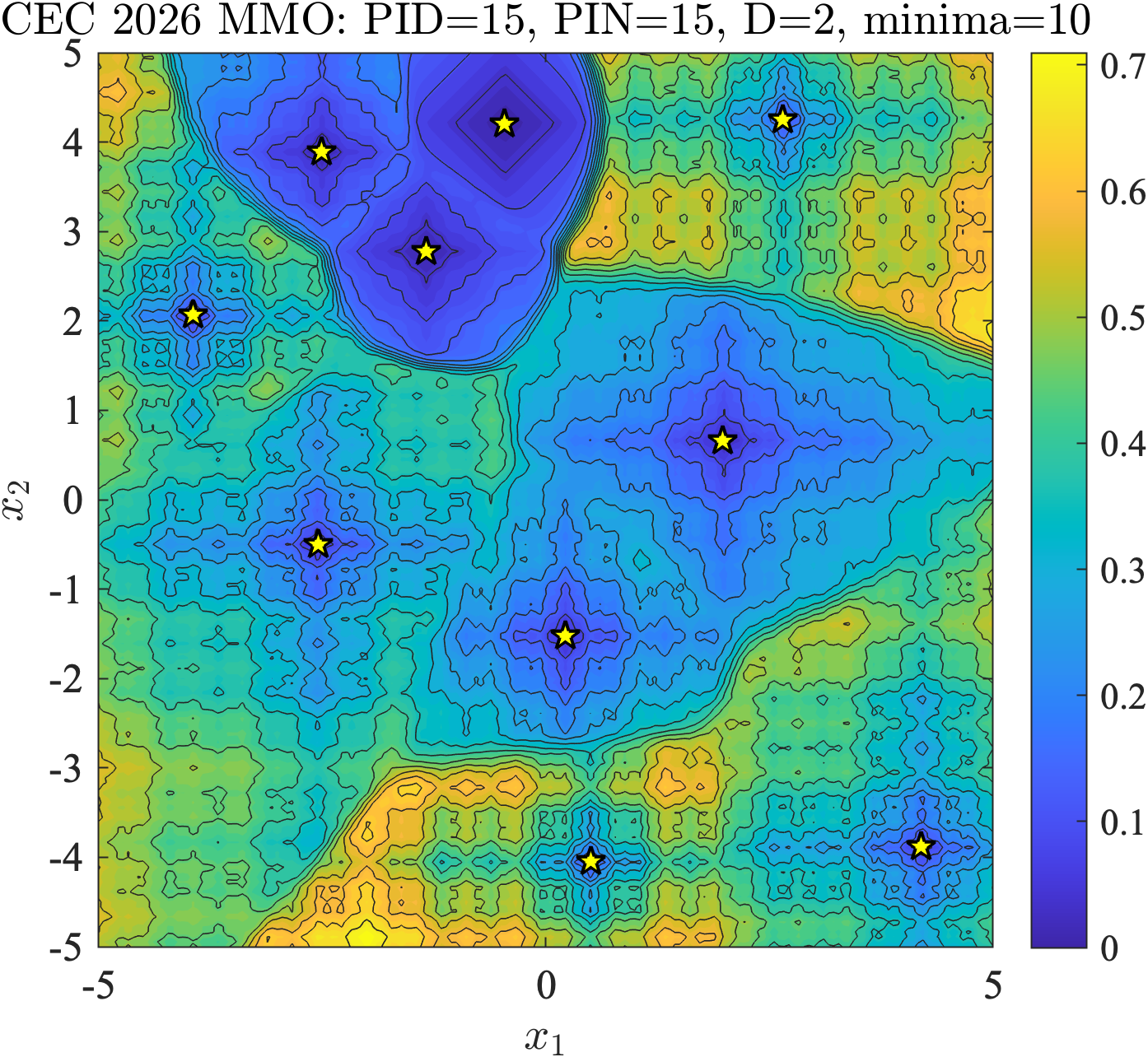}
\includegraphics[width=0.24\textwidth]{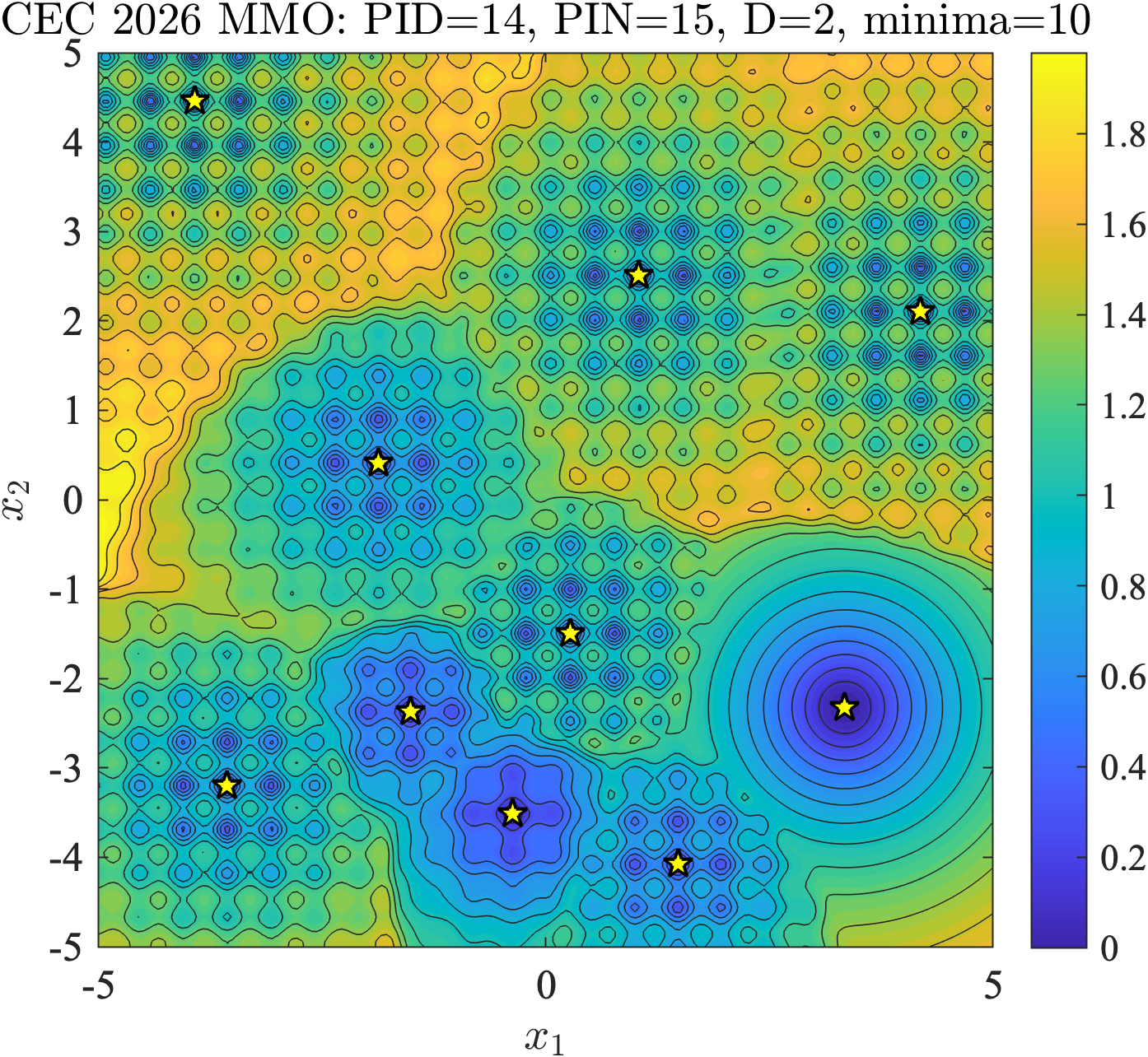}
\caption{Representative two-dimensional CEC 2026 MMO landscapes. The contour plots illustrate variation in basin size, basin shape, spatial distribution, local ruggedness, and orientation across selected PIDs and instances. Star markers indicate known global minimizers and are shown only for visualization; they are not used during optimization or final reporting.}
\label{fig:cec2026_landscapes}
\end{figure*}

The benchmark is designed to test both multimodal peak coverage and the ability to handle structural interactions among variables. Each problem instance may contain basins with different shapes, sizes, orientations, and spatial distributions. In addition, the variable dependencies are controlled through dependency blocks. Early instances contain stronger variable interaction, whereas later instances are closer to separable structures. Therefore, an effective algorithm must not only discover multiple optima but also adapt to different levels of variable coupling and basin geometry. Fig.~\ref{fig:cec2026_landscapes} illustrates these characteristics for selected two-dimensional cases.

\subsection{CEC 2026 Scoring Rule}
Unlike a peak-ratio-only assessment, the CEC 2026 score considers both coverage and reporting precision ~\cite{ahrari2026tr}. Let $\NGM$ denote the number of known global minimizers for a given problem instance and $\Nsol$ denote the number of solutions reported by an algorithm. The robust peak ratio, denoted by $\RPR$, measures the fraction of global minimizers successfully detected under the benchmark tolerance settings. In this setting, recall is equivalent to $\RPR$. The precision term accounts for the number of reported solutions. It is calculated as
\begin{equation}
\mathrm{precision}
=
\RPR \times \frac{\NGM}{\Nsol}.
\label{eq:precision}
\end{equation}
The F1-score is then computed as
\begin{equation}
\Fone
=
\frac{
2\cdot \mathrm{precision}\cdot \RPR
}{
\mathrm{precision}+\RPR
},
\label{eq:f1}
\end{equation}
where standard numerical safeguards are used when the denominator is zero. Finally, the run-level score is defined as
\begin{equation}
\Score
=
\frac{1}{2}(\RPR+\Fone).
\label{eq:score}
\end{equation}

Eqs~\eqref{eq:precision}-\eqref{eq:score} show that the final score rewards both successful peak detection and compact reporting. Reporting more candidates can increase the chance of covering additional optima, but it can also reduce precision when many reported candidates are redundant or do not correspond to distinct global minima. Therefore, the final solution set must balance high coverage and controlled report size. This scoring structure is one of the main motivations for the score-aware reporting mechanism proposed in this work.

\subsection{Overview of RS-CMSA-ESII}
RS-CMSA-ESII is an improved version of the covariance matrix self-adaptation evolution strategy with repelling subpopulations, originally developed for multimodal optimization~\cite{ahrari2017rscmsa,ahrari2022rscmsa2}. The method combines covariance-adaptive local search with archive-based repulsion. Its core idea is to repeatedly search promising basins while discouraging future searches from converging to already-identified optima.

The original RS-CMSA-ES uses repelling subpopulations, where each subpopulation has its own mean vector, step size, covariance matrix, and best solution. Inferior subpopulations are repelled from archived optima and from superior subpopulation centers. RS-CMSA-ESII simplifies and improves this structure by using one active subpopulation per restart. This design reduces algorithmic complexity and focuses the repulsion mechanism on the archived solutions. The active subpopulation evolves according to a CMSA-ES-style mechanism, where the sampling distribution is controlled by a mean vector, global step size, and covariance matrix.

Several improvements distinguish RS-CMSA-ESII from the earlier RS-CMSA-ES. These include adaptive normalized taboo-distance updates, an improved covariance update involving elite solutions, additional termination criteria for inefficient subpopulations, improved bound handling, more efficient initialization, and a refined estimate of critical taboo regions~\cite{ahrari2022rscmsa2}. These components are central to the robustness of RS-CMSA-ESII and are therefore preserved in the proposed method.

\subsection{Primary Archive, Taboo Regions, and Restarts}
The primary archive is the central memory component of RS-CMSA-ESII. Let
\begin{equation}\label{eq:primery_archive}
\Aset =
\left\{(x_m^A,f_m^A,\hat d_m^A)\right\}_{m=1}^{|\Aset|},
\end{equation}
denote the primary archive, where $x_m^A$ is an archived solution, $f_m^A$ is its objective value, and $\hat d_m^A$ is the normalized taboo distance associated with the archived basin. The normalized taboo distance controls the relative size of the taboo region around the archived solution.

During a restart, the active subpopulation samples new candidate solutions using its current covariance-adaptive mutation profile. Candidate solutions that fall inside critical taboo regions are rejected without evaluation. The taboo regions are centered at archived solutions and are shaped by the covariance matrix of the active subpopulation. Thus, the method can represent non-spherical exclusion regions, which is important for basins with different orientations and condition numbers.

At a high level, one restart of RS-CMSA-ESII consists of three main phases:
\begin{enumerate}
\item initialize a covariance-adaptive subpopulation;
\item evolve the subpopulation while respecting taboo-region rejection;
\item analyze the restart-best solution after termination.
\end{enumerate}
When the restart terminates, the best solution of that restart is compared with the current primary archive. Objective-value tests and hill-valley basin checks are used to decide whether the solution represents a new desirable basin or corresponds to an already archived optimum. If it represents a new basin, it is added to the primary archive. If it corresponds to an existing archived solution, the associated normalized taboo distance may be enlarged to reduce the probability that later restarts converge to the same basin. If the restart is unsuccessful or converges to an undesirable region, the taboo distances may be adjusted to avoid overly restrictive behavior in the archive.

This restart-and-archive mechanism gives RS-CMSA-ESII two important properties. First, it can learn from previously identified optima, reducing repeated convergence to the same basins. Second, because the covariance matrix adapts during each restart, the search can exploit local basin geometry. However, the original final reported set is primarily determined by the primary archive. Therefore, restart-level candidates that were visited during the search but not accepted into the primary archive may be lost. The proposed S-CARD-CMSA addresses this limitation by leveraging a passive secondary candidate archive and a score-aware, density-filtered rule, as described in the next section.

\section{Proposed Method: S-CARD-CMSA}\label{sec:proposed}

\subsection{Overall Framework}
This section presents S-CARD-CMSA, a score-aware candidate-archive framework with density-filtered reporting built on RS-CMSA-ESII. The method is developed as a conservative extension of RS-CMSA-ESII. The main design principle is to preserve the well-tested search engine of RS-CMSA-ESII and improve the candidate-retention and final-reporting stages, which are directly related to the CEC 2026 RPR-F1 scoring rule.


The overall framework consists of three stages. First, the original RS-CMSA-ESII search is executed using its primary archive and restart mechanism. Second, during each restart, a passive secondary candidate archive records the restart-best solution. Third, after the evaluation budget is exhausted, a score-aware density-filtered reporting rule constructs the final reported solution set from the union of the primary archive and filtered secondary candidates. The aim is to recover useful candidates that were visited during the search but not retained by the primary archive, while avoiding excessive or redundant reporting.

\subsection{Passive Secondary Candidate Archive}
In the original RS-CMSA-ESII reporting mechanism, the final reported set is mainly determined by the primary archive:
\begin{equation}
\Rset_{\mathrm{base}} = \Aset,
\label{eq:base_report}
\end{equation}where $\Aset$, defined in Eq. \eqref{eq:primery_archive}, denotes the primary archive maintained by RS-CMSA-ESII. However, a restart can visit a useful candidate without that candidate being accepted into the primary archive. This may happen when the candidate fails the archive's novelty, desirability, or basin-verification checks, even though it may still be close to a global minimizer under the final benchmark tolerance. Such candidates are lost if only the primary archive is reported.

To reduce this loss, S-CARD-CMSA maintains a passive secondary candidate archive:
\begin{equation}
\Cset =
\left\{(x_r^{\mathrm{best}},f_r^{\mathrm{best}},\eta_r)\right\}_{r=1}^{N_{\mathrm{restart}}},
\label{eq:secondary_archive}
\end{equation}
where $x_r^{\mathrm{best}}$ is the best candidate found in restart $r$, $f_r^{\mathrm{best}}$ is its objective value, and $\eta_r$ stores diagnostic information such as restart index, termination type, and evaluation count. The secondary archive is passive.
It only stores additional candidate information for final reporting.

Before final reporting, infeasible, non-finite, or clearly invalid secondary candidates are removed. The feasible and finite secondary archive is denoted by $\Cset_f$. The final candidate pool is then
\begin{equation}
\Pset = \Aset \cup \Cset_f.
\label{eq:pool}
\end{equation}

\subsection{Score-Aware Density-Filtered Reporting}
The final reporting rule maps the candidate pool $\Pset$ into the submitted solution set $\Rset$. The goal is to retain useful coverage from the secondary archive while controlling the number of reported candidates. This is important because the CEC 2026 score combines RPR and F1-score. A larger set of reported candidates may increase the chance of detecting more optima, but redundant or low-quality candidates can reduce precision and, in turn, the F1-score.

The proposed reporting rule has four steps: objective-value filtering, primary-archive preservation, near-duplicate handling, and density-filtered secondary insertion.

\subsubsection{Objective-Value Filtering}

Let
\begin{equation}
f_{\min} = \min_{x\in\Pset} f(x)
\end{equation}
be the best objective value in the candidate pool. Secondary candidates with clearly inferior objective values are discarded using a fixed value-window condition:
\begin{equation}\label{eq:value_window}
\Cset_v = \{x \in \Cset_f \mid f(x) \leq f_{\min}+\Delta_f\},
\end{equation}
where $\Delta_f$ is a globally fixed tolerance. This step prevents weak restart-level candidates from being included in the final report.

\subsubsection{Primary-Archive Preservation}
The primary archive contains candidates that passed the original RS-CMSA-ESII archive logic. These candidates are considered high-confidence. Therefore, the reporting rule first inserts primary archive members into $\Rset$, subject only to numerical validity and duplicate removal. This ensures that the proposed extension does not discard the main output of the base optimizer.

\subsubsection{Normalized Distance for Duplicate Control}

Candidate similarity is measured using normalized Euclidean distance:
\begin{equation}
\Dnorm(x_i,x_j)
=
\left\lVert
\frac{x_i-x_j}{u-l}
\right\rVert_2,
\label{eq:dnorm}
\end{equation}
where $u$ and $l$ are the upper and lower bound vectors. Normalization makes the distance measure independent of the absolute scale of the search space.

If two candidates are extremely close according to $\Dnorm$, they are treated as duplicate or near-duplicate representatives of the same region. In such a case, the candidate with the better objective value is retained.

\subsubsection{Density-Filtered Secondary Insertion}
After primary candidates are inserted, secondary candidates are considered in ascending order of objective value. For a secondary candidate $x$, its distance to the current reported set is
\begin{equation}
\rho(x)
=
\min_{y\in\Rset}
\Dnorm(x,y).
\label{eq:rho}
\end{equation}
The candidate is inserted into $\Rset$ if it is sufficiently separated from all currently reported candidates:
\begin{equation}
\rho(x) \geq \tau_{\rho},
\label{eq:density_threshold}
\end{equation}
where the density threshold is defined as $\tau_{\rho}(D)=\alpha_{\rho}\sqrt{D}.$ 
Here, $\alpha_{\rho}$ is a fixed density-filtering coefficient. If $x$ is close to an already reported candidate, the better objective-value representative is retained. This density-filtered rule reduces local redundancy in the final solution set and improves precision without strongly reducing coverage. In the S-CARD-CMSA setting, the value-window threshold and density coefficient are fixed as $\Delta_f=2\times10^{-3},~ \alpha_{\rho}=2\times10^{-3},$ for all problems, instances, and dimensions (See Subsection \ref{subsec:DF}). 

\subsection{Algorithmic Description}

Algorithm~\ref{alg:sca} summarizes the overall S-CARD-CMSA framework. Algorithm~\ref{alg:density_reporting} details the final density-filtered reporting procedure.

\begin{algorithm}[h!]
\caption{S-CARD-CMSA}
\label{alg:sca}
\begin{algorithmic}[1]
\STATE Initialize the RS-CMSA-ESII state and primary archive $\Aset$.
\STATE Initialize the passive secondary archive $\Cset=\emptyset$.
\WHILE{the evaluation budget is not exhausted}
\STATE Initialize one RS-CMSA-ESII restart using the original initialization rule.
\STATE Evolve the CMSA-ES subpopulation using the original RS-CMSA-ESII sampling, selection, covariance adaptation, taboo-region handling, and termination criteria.
\STATE Let $(x_r^{\mathrm{best}},f_r^{\mathrm{best}})$ be the best solution found during restart $r$.
\STATE Store $(x_r^{\mathrm{best}},f_r^{\mathrm{best}},\eta_r)$ in the passive secondary archive $\Cset$.
\STATE Update the primary archive $\Aset$ using the original RS-CMSA-ESII archive and hill-valley checking logic.
\STATE Update normalized taboo distances and restart state using the original RS-CMSA-ESII rules.
\ENDWHILE
\STATE Remove invalid candidates from $\Cset$ to obtain $\Cset_f$.
\STATE Form the candidate pool $\Pset=\Aset\cup\Cset_f$.
\STATE Apply score-aware density-filtered reporting to obtain $\Rset$ using Algorithm~\ref{alg:density_reporting}.
\STATE \textbf{return} final reported set $\Rset$.
\end{algorithmic}
\end{algorithm}

\begin{algorithm}[h!]
\caption{Score-aware density-filtered reporting}
\label{alg:density_reporting}
\begin{algorithmic}[1]
\REQUIRE Primary archive $\Aset$, feasible secondary archive $\Cset_f$, bounds $l,u$, value window $\Delta_f$, density threshold $\tau_{\rho}$.
\ENSURE Final reported solution set $\Rset$.
\STATE Initialize $\Rset=\emptyset$.
\STATE Form $\Pset=\Aset\cup\Cset_f$ and compute $f_{\min}=\min_{x\in\Pset}f(x)$.
\STATE Define $\Cset_v=\{x\in\Cset_f \mid f(x)\leq f_{\min}+\Delta_f\}$.
\STATE Sort primary archive candidates by objective value.
\FOR{each primary candidate $x\in\Aset$}
\STATE Insert $x$ into $\Rset$ if it is finite, feasible, and not an exact duplicate of an existing reported candidate.
\ENDFOR
\STATE Sort candidates in $\Cset_v$ by objective value.
\FOR{each secondary candidate $x\in\Cset_v$}
\STATE Compute $\rho(x)=\min_{y\in\Rset}\Dnorm(x,y)$.
\IF{$\Rset=\emptyset$ or $\rho(x)\geq\tau_{\rho}$}
\STATE Insert $x$ into $\Rset$.
\ELSE
\STATE Let $y^\star=\arg\min_{y\in\Rset}\Dnorm(x,y)$.
\IF{$f(x)<f(y^\star)$}
\STATE Replace $y^\star$ by $x$ only if this does not remove a protected primary candidate.
\ENDIF
\ENDIF
\ENDFOR
\STATE \textbf{return} $\Rset$.
\end{algorithmic}
\end{algorithm}

\subsection{Difference from Original RS-CMSA-ESII}

The proposed method differs from the original RS-CMSA-ESII only in candidate retention and final reporting. The original method reports primarily from the primary archive:
\begin{equation}\nonumber
\Rset_{\mathrm{base}}=\Aset.
\label{eq:r_base}
\end{equation}
In contrast, S-CARD-CMSA reports
\begin{equation}
\Rset_{\mathrm{SCA}}
=
\mathcal{D}_{\Delta_f,\tau_{\rho}}
\left(\Aset\cup\Cset_f\right),
\label{eq:r_sca}
\end{equation}
where $\mathcal{D}_{\Delta_f,\tau_{\rho}}(\cdot)$ denotes the score-aware density-filtered reporting operator. The search trajectory of the optimizer remains unchanged:
\begin{equation}\nonumber
\mathcal{S}_{\mathrm{SCA}} = \mathcal{S}_{\mathrm{RS\mbox{-}CMSA\mbox{-}ESII}},
\end{equation}
where $\mathcal{S}$ denotes the sequence of evaluated search points, up to negligible bookkeeping for storing restart-best candidates. Thus, any improvement arises from better use of already generated candidate information rather than from additional search evaluations or a modified sampling mechanism.

This distinction is important. The proposed method does not claim to redesign RS-CMSA-ESII. Instead, it addresses a competition-relevant limitation: useful restart-best candidates may be visited but not retained by the strict primary archive, while overly relaxed reporting can harm F1-score. S-CARD-CMSA balances these effects by combining passive candidate recovery with density-filtered precision control.

\section{Experimental Development and Ablation Study}\label{sec:experiments}

\subsection{Benchmark, Metrics, and Development Protocol}
Development experiments were conducted using the IEEE CEC 2026 multimodal optimization benchmark interface \cite{ahrari2026tr}. The benchmark provides the problem object, bounds, evaluation budget, and postprocessing routines. True global minima are not used during optimization. They are used only for offline development analysis to compute RPR, precision, F1-score, and score. Two development subsets were used.

The first subset was used for detailed reporting-rule comparison: $\mathrm{PID}\in\{1,2,6,7,10,11,14,15\},~
\mathrm{PIN}\in\{1,5,10,15\}$ for $D\in\{10,20\},~ \mathrm{run}=1,\ldots,5.$ This gives $8\times 4\times 2\times 5 = 320$ development runs. This subset emphasizes higher-dimensional and previously difficult cases.

The second subset was used for broader validation: $\mathrm{PID}=1,\ldots,16,~
\mathrm{PIN}\in\{1,5,10,15\},$ for $D\in\{2,5,10,20\},~ \mathrm{run}=1,2,3$ for a total of $16\times 4\times 4\times 3 = 768$ validation runs. Finally, a full submission-scale run was produced for all $16\times15\times4 = 960$ PID-PIN-dimension combinations. The evaluation focuses on the run-level score defined in Eq. \eqref{eq:score}. 
\begin{table*}[t]
\centering
\caption{Development variants considered in this work.}
\label{tab:variants}
\begin{tabularx}{\textwidth}{p{3.0cm}p{4.6cm}X p{3.0cm}}
\toprule
Variant & Description & Effect on search dynamics & Decision \\
\midrule
Primary-Archive Baseline & Original RS-CMSA-ESII reporting from the primary archive only & Original search and original primary-archive reporting & Baseline. \\
\midrule
SA-Strict & S-CARD-CMSA with passive secondary archive and strict cleanup & No change to sampling, covariance adaptation, taboo regions, or restarts & Accepted as an initial archive-enhanced variant; recovers useful restart-level candidates. \\
\midrule
SA-Relaxed & Passive secondary archive with relaxed final cleanup & No change to search; reports more secondary candidates & Improved RPR in some cases, but often over-reported and reduced precision/F1-score. \\
\midrule
Search-Modified SCA & Restart scheduling, adaptive final polishing, or covariance-mode variants & Changes search behavior or consumes final-refinement budget & Not retained; gains were inconsistent or negligible. \\
\midrule
SCA-Medium & Passive secondary archive with medium score-aware final reporting & No change to search; modifies only final reporting & Accepted as a strong safe reporting rule before density-filtered reporting. \\
\midrule
DimSCA & Dimension-aware reporting rule selection & No change to search; selects reporting rule based on dimension & Positive in some cases, but less stable than density-filtered reporting. \\
\midrule
NBNC-SCA & Nearest-better-neighbor-style final cleanup & No change to search & Rejected; pruning was too aggressive, reducing RPR. \\
\midrule
CapSCA & Score-aware candidate-count cap after final candidate pooling & No change to search & Rejected; precision improved, but excessive pruning reduced RPR. \\
\midrule
\textbf{DF-SCA (proposed S-CARD-CMSA)} & \textbf{Density-filtered score-aware reporting from the union of primary and secondary archives} & \textbf{No change to search; removes locally redundant final candidates} & \textbf{Selected; preserved RPR while improving precision and F1-score.} \\
\midrule
Archive-Aware Restart SCA & Archive-aware restart initialization variants & Changes restart initialization & Not retained; unstable across PIDs and did not improve the mean score. \\
\midrule
TLLS-SCA & Lightweight two-level local refinement after final reporting & Uses a small final-refinement budget & Not retained; score gain was negligible. \\
\midrule
CMAR-SCA & Covariance-shaped final candidate refinement inspired by CMAR & Uses a small final-refinement budget & Not retained; score gain was negligible. \\
\bottomrule
\end{tabularx}
\end{table*}
\subsection{Mathematical Description of Development Variants}
Let $\Aset$ denote the primary archive of RS-CMSA-ESII and let $\Cset_f$ denote the feasible and finite passive secondary candidate archive. The final candidate pool is defined in Eq. \eqref{eq:pool}. Let $\mathcal{V}_{\Delta_f}(\cdot)$ be an objective-value-window filter, $\mathcal{D}_{\tau_\rho}(\cdot)$ be a density-based duplicate-control operator, $\mathcal{N}_{\mathrm{NBNC}}(\cdot)$ be a nearest-better-neighbor-cleanup operator, and $\mathcal{K}_{N_{\max}}(\cdot)$ be a candidate-count cap operator. The development variants can be represented as follows.
\subsubsection{Candidate-Archive and Reporting Variants}\begin{enumerate}[(i)]
\item \textbf{Primary-Archive Baseline (Original RS-CMSA-ESII):} The baseline corresponds to the original RS-CMSA-ESII reporting mechanism, where only the primary archive is reported: $\Rset_{\mathrm{PA}} = \Aset.$
 \item \textbf{Strict Secondary-Archive Reporting (SA-Strict):}
This variant augments the primary archive with the passive secondary archive and applies a strict cleanup rule: $\Rset_{\mathrm{SA\mbox{-}Strict}}=
\mathcal{D}_{\tau_s}
\left(
\mathcal{V}_{\Delta_s}(\Aset\cup\Cset_f)
\right),$
where $\Delta_s$ and $\tau_s$ denote strict value and distance thresholds.

 \item \textbf{Relaxed Secondary-Archive Reporting (SA-Relaxed):}
A more relaxed version retains more secondary candidates: $\Rset_{\mathrm{SA\mbox{-}Relaxed}}=
\mathcal{D}_{\tau_r}
\left(
\mathcal{V}_{\Delta_r}(\Aset\cup\Cset_f)
\right),$ where $\Delta_r>\Delta_s$ or $\tau_r<\tau_s$ allows more candidates to be retained. This improves RPR in some cases but can reduce precision (see Subsection \ref{subsec:CA}).

 \item \textbf{Search-Modified SCA Variants (SearchMod):}
Several variants modified the search trajectory via restart scheduling, adaptive final polishing, or changes to covariance mode. Their generic final report can be written as $\Rset_{\mathrm{SearchMod}}=
\mathcal{D}
(
\Aset^{\prime}\cup\Cset_f^{\prime}),$ where $\Aset^{\prime}$ and $\Cset_f^{\prime}$ are produced by a modified search trajectory. These variants were not retained because their improvements were inconsistent or negligible.

 \item \textbf{Medium Score-Aware Reporting (SCA-Medium):}
This variant applies a medium-strength reporting rule designed to balance coverage and precision: $\Rset_{\mathrm{SCA\mbox{-}Medium}}=
\mathcal{D}_{\tau_m}
\left(
\mathcal{V}_{\Delta_m}(\Aset\cup\Cset_f)
\right),$ where $\Delta_m$ and $\tau_m$ are selected to balance coverage and precision. This was the strongest safe reporting rule before density-filtered reporting (see Subsection \ref{subsec:CA}). 
\end{enumerate}
\subsubsection{Advanced Final-Reporting Variants}
The following variants were designed to improve the final reporting stage while keeping the RS-CMSA-ESII search unchanged.
\begin{enumerate}[(i)]
\item \textbf{Dimension-Aware SCA Reporting (DimSCA):}
This variant selects a fixed reporting rule according to the problem dimension: $\Rset_{\mathrm{DimSCA}}=
\mathcal{R}_{g(D)}(\Aset,\Cset_f),$ where $g(D)$ maps each dimension to candidate reporting rules, such as Primary-Archive Baseline, SA-Strict, or SCA-Medium. Several fixed mappings were tested. For example, one mapping preserved SCA-Medium for $D=10$ and used SA-Strict for $D=20$, while another used SA-Strict for both tested dimensions. Since the 320-run subset contains only $D=10$ and $D=20$, the DimSCA results are reported as ranges over the tested mappings. Although this approach produced positive results in some cases, it was less stable than the selected density-filtered rule (see Subsection \ref{subsec:AFR}).

\item \textbf{NBNC-Guided Cleanup (NBNC-SCA):}
This variant applies a nearest-better-neighbor-style cleanup operator:
$
\Rset_{\mathrm{NBNC\mbox{-}SCA}}=
\mathcal{N}_{\mathrm{NBNC}}(\Aset\cup\Cset_f).$ This variant was motivated by nearest-better clustering ideas \cite{NBNC}, but in our final-candidate setting, it pruned too aggressively and reduced RPR (see Subsection \ref{subsec:AFR}).

\item \textbf{Capped Score-Aware Reporting (CapSCA):}
This variant applies a candidate-count cap after candidate pooling and cleanup:
$
\Rset_{\mathrm{CapSCA}}=
\mathcal{K}_{N_{\max}}
\left(
\mathcal{D}_{\tau}
(\mathcal{V}_{\Delta}(\Aset\cup\Cset_f))
\right),$ where $N_{\max}=
\max
\left(
|\Aset|+c,
\lceil \gamma |\Aset| \rceil
\right).$ Although this improved precision, it removed too many candidates and reduced RPR (see Subsection \ref{subsec:AFR}).

\item \textbf{Density-Filtered SCA Reporting (DF-SCA):}
This is the selected final reporting rule. It is defined as
$
\Rset_{\mathrm{DF\mbox{-}SCA}}=
\mathcal{D}_{\tau_{\rho}}
\left(
\mathcal{V}_{\Delta_f}(\Aset\cup\Cset_f)
\right).$ For a secondary candidate $x$, the local density distance is defined in Eq. \eqref{eq:rho}.
The candidate is inserted if $\rho(x)\geq \tau_{\rho}.$
If a candidate is close to an existing reported solution; the better objective-value representative is retained, while protected primary candidates are not removed by secondary candidates. This rule preserves coverage while reducing redundant reports (see Subsection \ref{subsec:AFR}).
\end{enumerate}

\subsection{Summary of Development Variants}
Table~\ref{tab:variants} summarizes the major variants considered during development. The final retained branch consists of the passive secondary archive and score-aware density-filtered reporting. The table uses descriptive method names rather than internal development labels, so the ablation study is directly linked to the proposed S-CARD-CMSA framework. Variants that changed the internal restart behavior or used additional final-refinement steps were also tested (Subsection \ref{subsec:rejected_variants}), but they were not retained because their improvements were inconsistent or negligible.

\subsection{Comparison of Candidate-Archive and Reporting Variants}\label{subsec:CA}
Table~\ref{tab:reporting_rules} compares the main reporting rules on the 320-run development subset. The Primary-Archive Baseline reports only the original RS-CMSA-ESII primary archive. SA-Strict adds the passive secondary archive but applies strict cleanup. SA-Relaxed retains more secondary candidates and therefore gives the highest mean RPR; however, its larger report size reduces precision and F1-score. SCA-Medium provides a better balance between coverage and precision. The selected DF-SCA rule further improves this balance by keeping the same mean RPR as SCA-Medium while improving precision, F1-score, and the final score.

\begin{table}[!h]
\centering
\caption{Comparison of reporting rules on the 320-run development subset.}
\label{tab:reporting_rules}
\resizebox{\linewidth}{!}{\begin{tabular}{lccccc}
\toprule
Rule & Mean RPR & Mean precision & Mean F1 & Mean score & Mean $\Nsol$ \\
\midrule
RS-CMSA-ESII & 0.5116 & \textbf{0.9926} & 0.6406 & 0.5761 &  7.89 \\
SA-Strict & 0.5545 & 0.8188 & 0.6370 & 0.5957 & 10.38 \\
SA-Relaxed & \textbf{0.5590} & 0.7438 & 0.6119 & 0.5854 & 11.67 \\
SCA-Medium & 0.5589 & 0.8516 & 0.6434 & 0.6012 & 10.14 \\
\textbf{DF-SCA} & 0.5589 & 0.8705 & \textbf{0.6509} & \textbf{0.6049} & 9.82 \\
\bottomrule
\end{tabular}}
\end{table}

Compared with SCA-Medium, the selected DF-SCA rule improves the mean score from 0.6012 to 0.6049. It also reduces the mean number of reported solutions from 10.14 to 9.82 while maintaining the same mean RPR. This confirms that the improvement is obtained through better precision control rather than simply reporting more candidates. In other words, DF-SCA preserves the coverage benefit of the secondary archive while removing locally redundant candidates that would otherwise reduce the F1-score.

\subsection{Comparison of Advanced Final-Reporting Variants}\label{subsec:AFR}
Table~\ref{tab:bseries} summarizes the advanced final-reporting variants evaluated on the 320-run development subset. These variants were designed after SCA-Medium was identified as a strong safe reporting baseline. All variants in this group keep the RS-CMSA-ESII search unchanged and differ only in the way the final candidate set is constructed from the primary and secondary archives.

\begin{table*}[!h]
\centering
\caption{Summary of advanced final-reporting variants on the 320-run development subset.}
\label{tab:bseries}
\resizebox{\linewidth}{!}{\begin{tabular}{lccccc}
\toprule
Variant & Mean RPR & Mean precision & Mean F1 & Mean score & Main observation \\
\midrule
SCA-Medium & 0.5589 & 0.8516 & 0.6434 & 0.6012 & Strong safe baseline. \\
DimSCA & 0.5439-0.5575 & 0.8546-0.9055 & 0.6474-0.6618 & 0.6024-0.6028 & Positive but less stable. \\
NBNC-SCA & Low & High & Low & Poor & Pruned too many candidates. \\
CapSCA & 0.5116 & \textbf{0.9969} & 0.6411 & 0.5763 & Too conservative; RPR loss. \\
\textbf{DF-SCA} & 0.5589 & 0.8705 & \textbf{0.6509} & \textbf{0.6049} & Best balance of RPR and precision. \\
\bottomrule
\multicolumn{6}{l}{\footnotesize Qualitative entries indicate that NBNC-SCA was clearly dominated due to severe RPR loss.}
\end{tabular}}
\end{table*}
The comparison shows that the main challenge is not simply to reduce the number of reported solutions. A very strict rule, such as CapSCA, can increase precision but may remove useful candidates and reduce RPR. Conversely, a relaxed rule can improve RPR but may over-report and reduce the F1-score. DimSCA provides small improvements in some settings by selecting different reporting rules for different dimensions, but its behavior is less stable. NBNC-SCA was motivated by nearest-better-neighbor clustering, but in the final reporting stage, it was too aggressive and removed candidates that contributed to peak coverage.

The selected DF-SCA rule provides the best compromise. It keeps the same mean RPR as SCA-Medium while increasing precision from 0.8516 to 0.8705 and F1-score from 0.6434 to 0.6509. The mean score improves from 0.6012 to 0.6049. This confirms that DF-SCA improves the final score by removing locally redundant candidates rather than by either over-reporting or excessively pruning the candidate set.

\subsection{Sensitivity to the Density-Filtering Coefficient}\label{subsec:DF}

The selected DF-SCA rule uses a normalized distance threshold to decide whether a secondary candidate is sufficiently separated from the currently reported set. To avoid problem-specific tuning while accounting for dimensional scaling, the threshold is defined as $\tau_{\rho}(D)=\alpha_{\rho}\sqrt{D},$
where $\alpha_{\rho}$ is a fixed density-filtering coefficient. Three coefficient settings were evaluated on the 320-run development subset: a low-density coefficient, a balanced coefficient, and a high-density coefficient. These variants are denoted as DF-SCA-Low, DF-SCA-Balanced, and DF-SCA-High, respectively.

Table~\ref{tab:density_threshold_sensitivity} reports the sensitivity results. DF-SCA-Low uses a smaller density coefficient and therefore allows more nearby candidates to remain in the final report. This limits the precision improvement. DF-SCA-High changes the filtering behavior more strongly, but it also reduces precision and F1-score in the tested subset. The intermediate setting, DF-SCA-Balanced, gives the best mean score by maintaining the same mean RPR as DF-SCA-Low while improving precision and F1-score. Therefore, $\alpha_{\rho}=0.002$ was selected as the final density-filtering coefficient.

\begin{table}[!h]
\centering
\caption{Sensitivity of DF-SCA to the density-filtering coefficient on the 320-run development subset.}
\label{tab:density_threshold_sensitivity}
\resizebox{\linewidth}{!}{
\begin{tabular}{lcccccc}
\toprule
Variant & $\alpha_{\rho}$ & Mean RPR & Mean precision & Mean F1 & Mean score & Mean $\Nsol$ \\
\midrule
DF-SCA-Low & 0.001 & 0.558904 & 0.849994 & 0.643001 & 0.600953 & 10.153 \\
\textbf{DF-SCA-Balanced} & \textbf{0.002} & 0.558904 & \textbf{0.870491} & \textbf{0.650922} & \textbf{0.604913} & \textbf{9.819} \\
DF-SCA-High & 0.005 & \textbf{0.558954} & 0.810685 & 0.637740 & 0.598347 & 10.494 \\
\bottomrule
\end{tabular}}
\end{table}

These results indicate that the final score is sensitive to the density-filtering coefficient. A very small or very large coefficient can retain less favorable candidate sets, either by allowing more local redundancy or by changing the replacement behavior in a way that reduces precision. The balanced coefficient, $\alpha_{\rho}=0.002$, provides the most reliable RPR-F1 trade-off and is therefore used in the final S-CARD-CMSA implementation.

\subsection{Broader 768-Run Validation}
After the 320-run development comparison, the selected DF-SCA rule was further validated on a broader 768-run subset. This subset included all 16 PIDs, four representative instances per PID, all four dimensions, and three independent runs. The purpose of this broader validation was to determine whether the improvement observed in the 320-run development subset remained stable across additional PIDs and lower-dimensional cases.

Table~\ref{tab:broad_validation} compares SCA-Medium and the selected DF-SCA rule on this 768-run subset. DF-SCA keeps the same mean RPR as SCA-Medium while improving precision and F1-score. The mean score improves from 0.816366 to 0.818328, corresponding to an absolute gain of 0.001963. The win/loss/tie count of DF-SCA relative to SCA-Medium is 84/6/678, indicating that the improvement is small but stable and rarely harmful.

\begin{table}[h!]
\centering
\caption{Broader validation of the selected DF-SCA reporting rule on 768 runs.}
\label{tab:broad_validation}
\resizebox{\linewidth}{!}{\begin{tabular}{lccccc}
\toprule
Rule & Mean RPR & Mean precision & Mean F1 & Mean score & Mean $\Nsol$ \\
\midrule
SCA-Medium &\bf 0.805386 & 0.907076 & 0.827346 & 0.816366 & 13.798 \\
\textbf{DF-SCA} & \textbf{0.805386} & \textbf{0.914775} & \textbf{0.831271} & \textbf{0.818328} & \textbf{13.589} \\
\bottomrule
\end{tabular}}
\end{table}

The improvement comes primarily from better precision control. DF-SCA reduces the mean number of reported solutions from 13.798 to 13.589 while maintaining the same mean RPR. This indicates that the density-filtered rule removes locally redundant candidates without sacrificing average peak coverage.

Dimension-wise, DF-SCA improves the mean score in all tested dimensions. The observed score gains are 0.00256 for $D=2$, 0.00101 for $D=5$, 0.00234 for $D=10$, and 0.00194 for $D=20$. Although these gains are modest, their consistency across dimensions supports selecting DF-SCA as the final reporting mechanism. The result also confirms the intended role of density filtering: it improves the RPR-F1 trade-off by reducing redundant reports rather than changing the underlying RS-CMSA-ESII search process.

\begin{figure*}[t]
\centering
\begin{minipage}{0.32\textwidth}
    \centering
    \includegraphics[width=\linewidth]{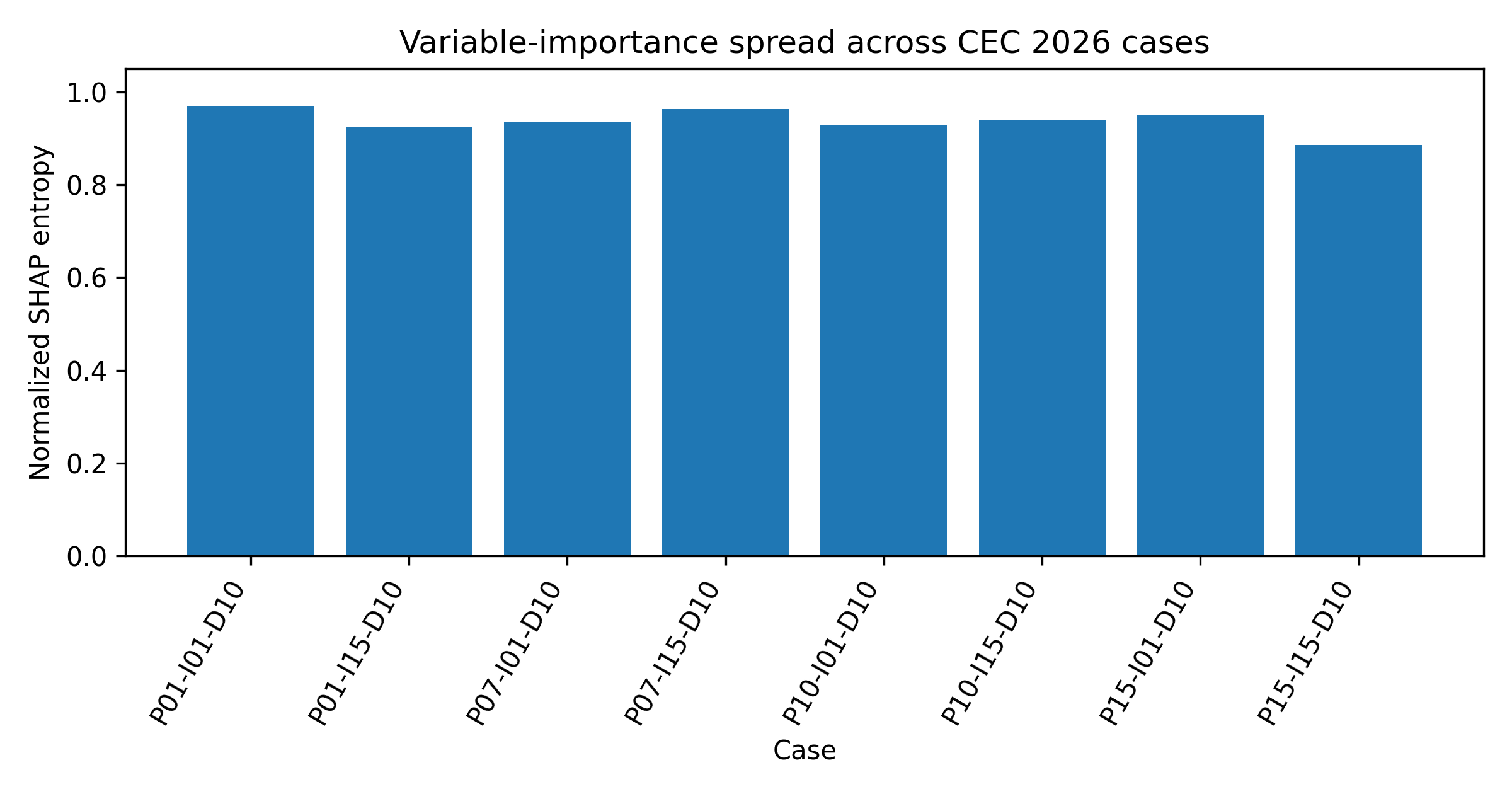}\\
   \footnotesize  (a) Normalized SHAP entropy
\end{minipage}
\hfill
\begin{minipage}{0.32\textwidth}
    \centering
    \includegraphics[width=\linewidth]{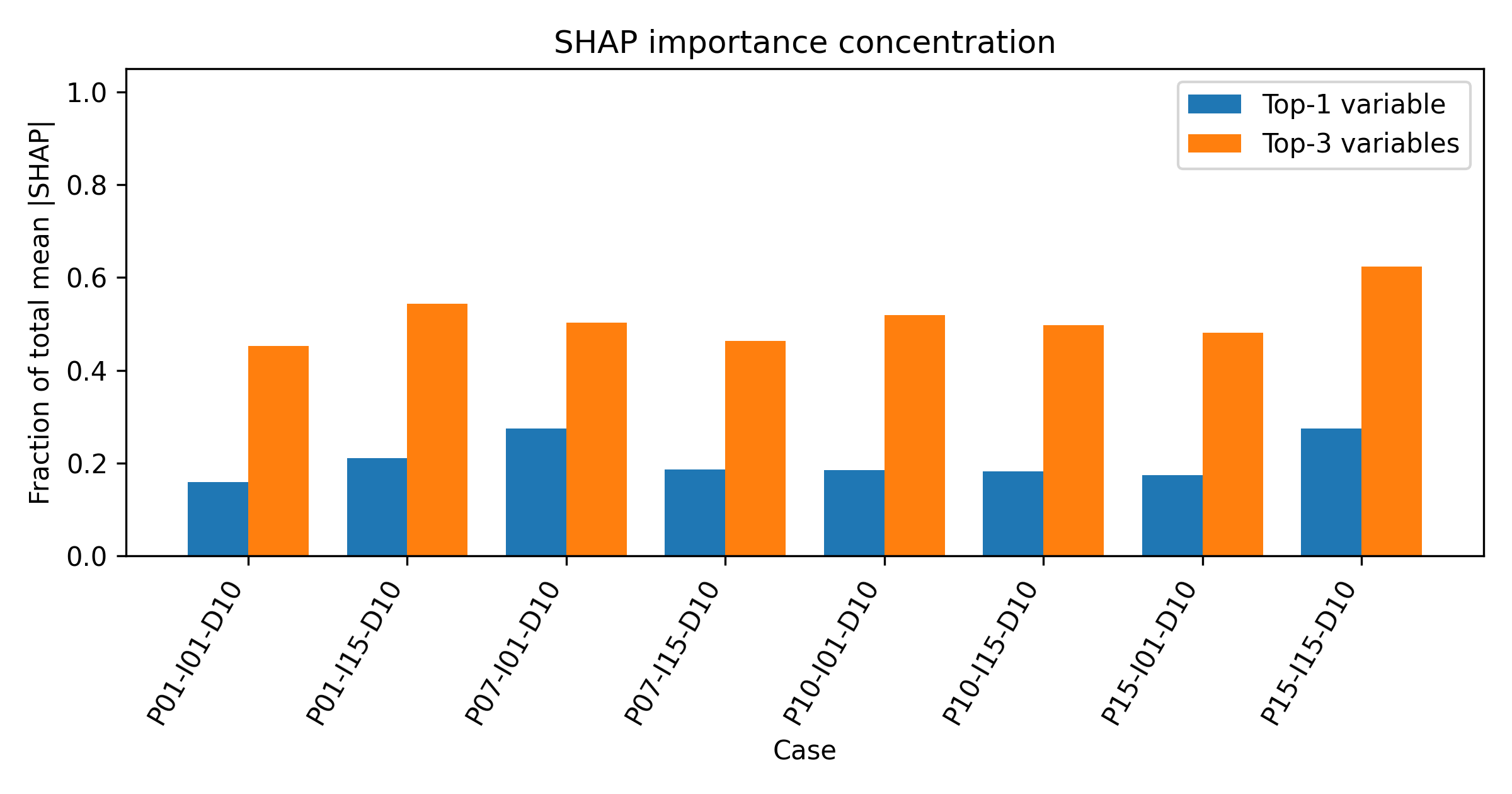}\\
    \footnotesize (b) Top-1 and top-3 concentrations
\end{minipage}
\hfill
\begin{minipage}{0.32\textwidth}
    \centering
    \includegraphics[width=\linewidth]{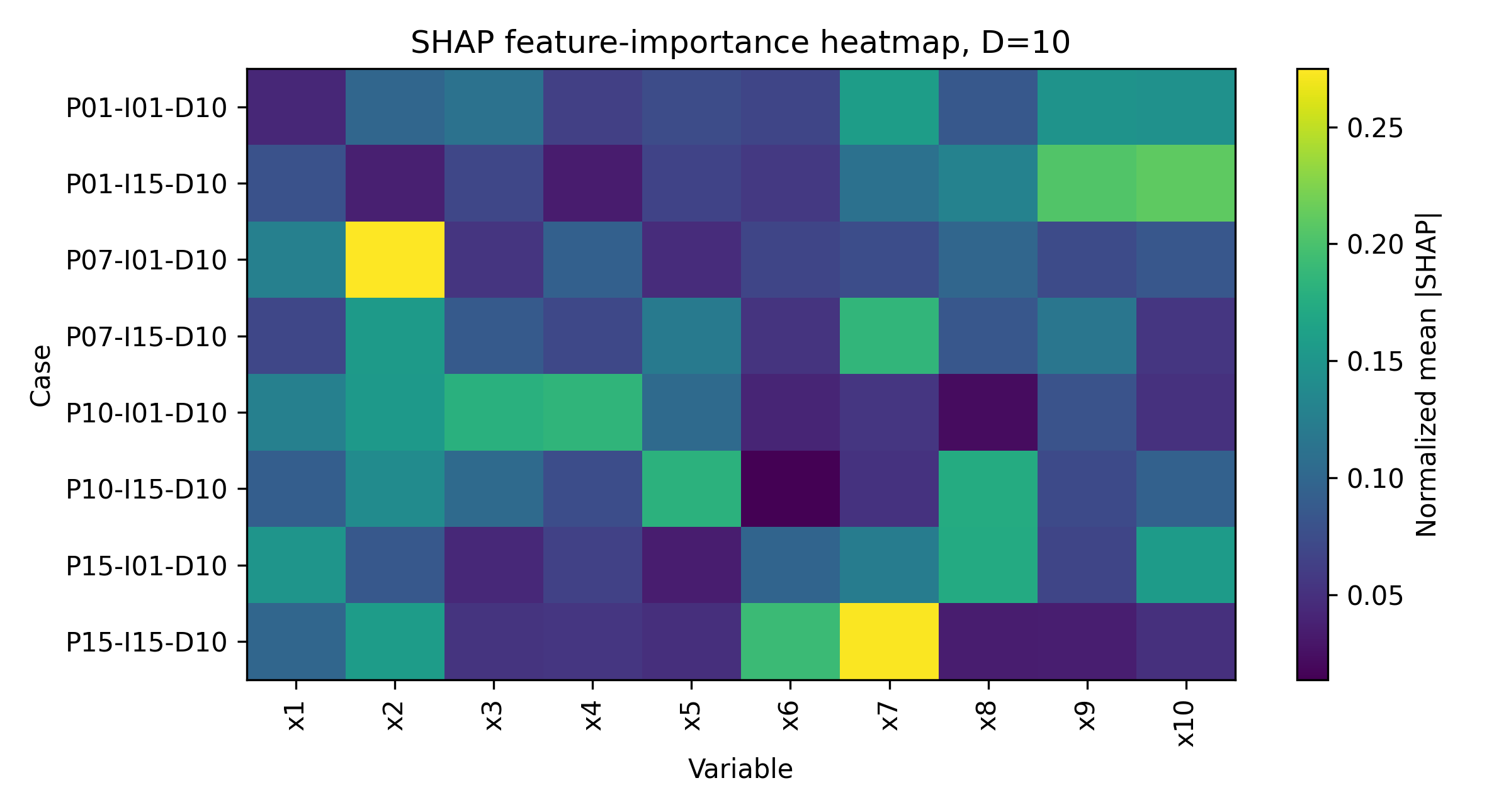}\\
   \footnotesize  (c) Normalized feature-importance heatmap
\end{minipage}
\caption{Post-hoc SHAP-based interpretation of selected CEC 2026 $D=10$ cases. The analysis is based on tree-surrogate models trained from sampled benchmark evaluations and is used only for landscape interpretation. High normalized entropy and moderate top-$k$ concentration indicate that the objective variation is distributed across several variables. The heatmap further shows that the dominant variables change across PIDs and instances.}
\label{fig:shap_landscape}
\end{figure*}
\subsection{Post-hoc Landscape Interpretation}
To further examine the structural difficulty of selected CEC 2026 cases, a post-hoc surrogate-based SHAP analysis was conducted. This analysis was used only for landscape interpretation and was not used during optimization, candidate selection, archive update, or final reporting. Eight representative $D=10$ cases were considered: $\mathrm{PID}\in\{1,7,10,15\},\quad \mathrm{PIN}\in\{1,15\}.$ For each case, $N=10{,}000$ points were sampled uniformly within the official search bounds and evaluated using the benchmark function. A tree-based surrogate model $\hat y$ was then trained to approximate a log-shifted objective value,
\begin{equation}
    y(x)=\log_{10}\left(1+f(x)-f_{\min}^{\mathrm{sample}}\right),
    \label{eq:shap_target}
\end{equation}
where $f_{\min}^{\mathrm{sample}}$ is the best sampled objective value for the corresponding case. SHAP values were then computed from the trained surrogate, not from the optimizer.

For a sampled point $x$, the surrogate prediction can be decomposed as
\begin{equation}
\hat y(x)
    =
    \phi_0+\sum_{j=1}^{D}\phi_j(x),
    \label{eq:shap_decomposition}
\end{equation}
where $\phi_0$ is the baseline prediction and $\phi_j(x)$ is the contribution of variable $x_j$ to the surrogate prediction. The global importance of variable $x_j$ is measured by the mean absolute SHAP value:
\begin{equation}
    I_j
    =
    \frac{1}{N_s}
    \sum_{i=1}^{N_s}
    \left|\phi_j(x^{(i)})\right|,
    \label{eq:shap_importance}
\end{equation}
where $N_s$ is the number of samples used for SHAP evaluation. To compare importance distributions across cases, the normalized importance is defined as:
\begin{equation}
    p_j
    =
    \frac{I_j}{\sum_{k=1}^{D} I_k}.
    \label{eq:shap_normalized}
\end{equation}
The spread of variable importance is summarized using normalized SHAP entropy:
\begin{equation}
    H_{\mathrm{SHAP}}
    =
    -
    \frac{1}{\log D}
    \sum_{j=1}^{D} p_j\log(p_j+\epsilon),
    \label{eq:shap_entropy}
\end{equation}
where $\epsilon$ is a small numerical constant. A high value of $H_{\mathrm{SHAP}}$ indicates that objective variation is distributed across many variables, whereas a low value indicates that a few variables dominate. In addition, the top-$k$ importance concentration is computed as
\begin{equation}
    C_k
    =
    \sum_{j\in \mathcal{T}_k} p_j,
    \label{eq:shap_topk}
\end{equation}
where $\mathcal{T}_k$ is the set of the $k$ variables with the largest normalized SHAP importance.

The surrogate models gave moderate predictive accuracy, with test $R^2$ values between $0.506$ and $0.691$ across the eight cases. Therefore, the SHAP results should be interpreted as approximate indicators of landscape structure rather than exact analytical decompositions of the benchmark functions. Nevertheless, the resulting patterns are informative. As shown in Fig.~\ref{fig:shap_landscape}, the normalized SHAP entropy is high for all selected cases, ranging from $0.886$ to $0.968$. This indicates that the objective variation is not dominated by a single variable; instead, many variables contribute to the sampled landscape. The top-1 variable accounts for only $15.9\%$-$27.5\%$ of the total mean absolute SHAP importance, while the top-3 variables account for $45.3\%$-$62.4\%$. This distributed importance pattern is consistent with the difficulty of multimodal coverage in higher-dimensional cases.

The heatmap in Fig.~\ref{fig:shap_landscape}(c) also shows that the dominant variables change across PIDs and instances. For example, the most influential variables are $(x_7,x_9,x_{10})$ for PID~1, PIN~1, but $(x_{10},x_9,x_8)$ for PID~1, PIN~15. Similarly, PID~10 shifts from $(x_4,x_3,x_2)$ in PIN~1 to $(x_5,x_8,x_2)$ in PIN~15, while PID~15 shifts from $(x_8,x_{10},x_1)$ to $(x_7,x_6,x_2)$. These changes indicate that variable relevance is instance-dependent, supporting the benchmark design goal of testing algorithms under different dependency and landscape structures. This observation also helps explain why simple fixed reporting or restart rules may not be uniformly effective across all cases.

\subsection{Rejected Internal and Local-Refinement Variants}\label{subsec:rejected_variants}
After selecting DF-SCA, several additional variants were tested to determine whether internal search changes or final local refinement could further improve the results. These included archive-aware restart initialization, TLLS-SCA, and CMAR-SCA. Although some individual runs improved, none of these variants produced a consistent mean-score improvement over DF-SCA. Therefore, they were not retained in the final method.

\subsubsection{Archive-Aware Restart Initialization}

The archive-aware restart (AAR) variants attempted to initialize future restarts away from already stored candidates. In generic form, a new restart center was chosen from random candidate centers $\{z_k\}_{k=1}^{K}$ by maximizing the distance to a memory set $\Mset$:
\begin{equation}
z^\star=\arg\max_{z_k}
\min_{y\in\Mset}
\left\lVert
\frac{z_k-y}{u-l}
\right\rVert_2.
\label{eq:a8_restart}
\end{equation}
Different memory sets and activation rules were tested:
\begin{equation}
\Mset =
\Aset\cup\Cset_f
\quad \text{or} \quad
\Mset=\Aset.
\end{equation}
The tested variants included full AAR, $D=20$-only activation, weak activation probability, and primary-archive-only restart repulsion.

Table~\ref{tab:a8} summarizes the results on the 120-run internal-change subset: $\mathrm{PID}\in\{1,6,7,14,15\},~
\mathrm{PIN}\in\{1,5,10,15\},~
D\in\{10,20\}.$ None of the variants improved over the DF-SCA baseline.

\begin{table}[h!]
\centering
\caption{Archive-aware restart variants compared with the DF-SCA baseline on 120 runs.}
\label{tab:a8}
\resizebox{\linewidth}{!}{\begin{tabular}{p{3cm}ccccc}
\toprule
Variant & Mean score & $\Delta$ score vs. baseline & Wins & Losses & Ties \\
\midrule
\textbf{DF-SCA} (baseline) & \textbf{0.564970} & -- & -- & -- & -- \\
AAR-SCA & 0.561122 & -0.003848 & 41 & 46 & 33 \\
AAR-SCA-$D20$ & 0.564218 & -0.000752 & 20 & 23 & 77 \\
Weak AAR-SCA & 0.563285 & -0.001685 & 44 & 53 & 23 \\
Primary-only AAR-SCA with repulsion & 0.563688 & -0.001281 & 37 & 49 & 34 \\
\bottomrule
\end{tabular}}
\end{table}

These results indicate that starting far from archived candidates does not necessarily place a restart near an undiscovered global basin. In some cases, it moves the search away from partially explored but still useful regions. The best AAR variant was still slightly worse than DF-SCA, so internal restart modification was not retained.

\subsubsection{TLLS-Lite Final Refinement}

A lightweight two-level local-search variant \cite{ANDE} was tested after density-filtered reporting. For a selected candidate $x$, local samples were generated as
\begin{equation}
x' = x + \sigma (u-l)\odot z,\quad z\sim\mathcal{N}(0,I),
\label{eq:tlls}
\end{equation}
with a small budget and candidate replacement only if improvement occurred. Let $\Phi_{\mathrm{TLLS}}$ denote this refinement operator. The final set is
\begin{equation}
\Rset_{\mathrm{TLLS}}=\mathcal{D}_{\tau_{\rho}}
\left(
\Phi_{\mathrm{TLLS}}(\Rset_{\mathrm{DF\mbox{-}SCA}})
\right).
\end{equation}
This variant was safe but produced an average score gain of only approximately $2.5\times10^{-5}$ on the 120-run test. Therefore, it was not retained.

\subsubsection{CMAR-Lite Final Refinement}
A covariance-matrix-adapted local refinement, denoted CMAR-SCA, was inspired by the Covariance Matrix Adapted Retreat phase used in EBOwithCMAR \cite{EBOwithCMAR}. For a selected candidate $x$, local candidates were sampled as
\begin{equation}
x' = x + \sigma B D z,
\label{eq:cmar}
\end{equation}
where $BD$ represents a covariance-shaped sampling matrix and $z$ is an isotropic random vector. The refined candidate replaced the original one only if it improved the objective value. Let $\Phi_{\mathrm{CMAR}}$ denote this refinement operator:
\begin{equation}
\Rset_{\mathrm{CMAR}}=\mathcal{D}_{\tau_{\rho}}
\left(
\Phi_{\mathrm{CMAR}}(\Rset_{\mathrm{DF\mbox{-}SCA}})
\right).
\end{equation}
This variant also remained safe, but its average score gain was only approximately $1.5\times10^{-5}$. Since the improvement was negligible, it was not retained.

\subsection{Reason for Selecting the Final Version}
The final selected method is S-CARD-CMSA with density-filtered reporting, denoted DF-SCA. This version was selected because it offered the best balance between score improvement, stability, and implementation risk. It preserves the original RS-CMSA-ESII search process and improves only the final use of candidate information through a passive secondary archive and density-filtered reporting. This allows useful restart-best candidates to be reconsidered without consuming additional function evaluations or altering the base optimizer.

The experimental results show that DF-SCA directly supports the CEC 2026 RPR-F1 scoring structure. It maintains the peak coverage obtained by SCA-Medium while reducing locally redundant reports, thereby improving precision, F1-score, and the overall score. In contrast, AAR variants were unstable across PIDs, and TLLS-SCA and CMAR-SCA produced only negligible score gains. Therefore, DF-SCA was selected as the final version, while SCA-Medium was retained as a safe backup during development. 

The final selected rule was also used to generate the complete set of $16\times15\times4=960$ submission files, and the generated files were checked for completeness, finite values, correct dimensionality, and bound feasibility.

\section{Discussion}\label{sec:discussion}




\subsection{Strengths of the Proposed Method}
The main strength of S-CARD-CMSA is that it improves final candidate reporting while preserving the robust search behavior of RS-CMSA-ESII. The passive secondary archive stores restart-best candidates that may otherwise be discarded, and DF-SCA filters the combined candidate set to reduce local redundancy. This improves the RPR-F1 trade-off without changing the underlying optimization process.

Another advantage is that the proposed extension is low-risk and reproducible. It does not require problem-specific tuning, does not use true global minimizers during optimization, and does not consume additional function evaluations for search. Therefore, the improvement comes from better use of information already generated by the base optimizer.

\subsection{Limitations}
The proposed method is primarily a candidate-retention and reporting extension. It does not fundamentally change the ability of RS-CMSA-ESII to discover new basins. Therefore, if the base search process never reaches a global basin, the final reporting mechanism cannot recover it.

Another limitation is that the density-filtering coefficient is fixed globally. Although the actual threshold is dimension-scaled, $\tau_{\rho}(D)=\alpha_{\rho}\sqrt{D}$, the coefficient $\alpha_{\rho}$ is not adapted to individual problems, instances, or landscape structures. This improves simplicity and avoids problem-specific tuning, but some cases may benefit from different thresholds depending on basin spacing, dimension, or modality structure. Dimension-aware or landscape-aware reporting may provide further gains, but the current development results did not justify replacing the fixed-coefficient density-filtered rule.

The tested internal restart variants also showed that archive-aware exploration is not automatically beneficial. Moving new restarts away from archived candidates may help on some PIDs but hurt others. This suggests that more reliable landscape-aware restart mechanisms would be needed before modifying the internal restart behavior.

\subsection{Why Internal Changes Were Not Retained}

Several internal or refinement-level modifications were tested after the density-filtered rule was selected. AAR initialization attempted to improve basin discovery by starting future restarts far from archived candidates. However, this strategy was unstable across PIDs and did not improve the mean score. TLLS-Lite and CMAR-Lite improved some objective values locally, but their effects on RPR, F1-score, and overall score were negligible.

These results suggest that the current performance gap is not primarily due to insufficient final local accuracy. Instead, the main remaining difficulty is discovering basins within a limited budget. Simple restart repulsion or final local refinement is not sufficient to robustly solve this issue.

\subsection{Future Work}

Future work can extend S-CARD-CMSA in several directions. First, adaptive reporting thresholds could be developed using measurable properties of the final candidate set, such as local candidate density, objective-value distribution, and archive stability. Second, landscape-aware restart strategies could be designed more carefully, using evidence of unexplored regions rather than only distance from archived candidates. Third, clustering-based basin grouping, such as nearest-better or hill-valley-assisted grouping, could be integrated more conservatively into the final reporting stage. Finally, a multi-stage framework could combine RS-CMSA-ESII's covariance-adaptive exploitation with a separate lightweight exploration component, provided that the evaluation budget and reporting precision remain controlled.

Overall, the development results indicate that score-aware candidate retention and reporting are effective, low-risk extensions for the CEC 2026 benchmark, while deeper search-level modifications require more extensive design and testing.

\section{Conclusion}\label{sec:conclusion}
This paper presented S-CARD-CMSA, a score-aware candidate-archive extension of RS-CMSA-ESII for the IEEE CEC 2026 multimodal optimization competition. The method preserves the core RS-CMSA-ESII search dynamics and introduces two conservative extensions: a passive secondary candidate archive and a score-aware density-filtered final reporting rule. Development experiments show that the secondary archive improves coverage by recovering restart-level candidates that the primary archive did not retain, while density filtering improves precision and F1-score by suppressing locally redundant candidates. Search-level extensions, including restart scheduling, local polishing, and covariance-mode switching, were tested but not retained because their improvements were inconsistent or negligible. The final method therefore emphasizes a robust and reproducible archive/reporting-level enhancement of a strong existing MMO optimizer.

\section*{Acknowledgment}
The author thanks the organizers of the IEEE CEC 2026 Competition on Benchmarking Niching Methods for Multimodal Optimization for providing the benchmark suite and experimental setup. The author also acknowledges the original developers of RS-CMSA-ES and RS-CMSA-ESII, on which the proposed framework is built.

\bibliographystyle{ieeetr}
\bibliography{sca_rs_cmsa_esii_refs}

\end{document}